\pdfoutput=1

\documentclass[11pt]{article}

\usepackage{emnlp2021}
\usepackage{times}
\usepackage{graphicx}
\usepackage{latexsym}
\usepackage{booktabs} 
\usepackage{subfigure}
\usepackage{bbding}
\usepackage{pifont}
\usepackage{multirow}
\usepackage{color}
\usepackage{amsmath}
\usepackage[T1]{fontenc}

\usepackage[utf8]{inputenc}
\usepackage{microtype}
\usepackage{amssymb}
\title{Knowledge Enhanced Fine-Tuning for Better Handling Unseen Entities in Dialogue Generation}

\author{
  Leyang Cui$^{\heartsuit \spadesuit}$\thanks{\ \  Contribution during internship at MSRA.}, Yu Wu$^\Diamond$, Shujie Liu$^\Diamond$, Yue Zhang$^{\spadesuit}$ \\
  $^\heartsuit$Zhejiang University \\
  $^\spadesuit$Westlake University \\
  $^\Diamond$Microsoft Research Asia \\
  \{cuileyang,zhangyue\}@westlake.edu.cn\  
  \{Wu.Yu,shujliu\}@microsoft.com \\
  }

\begin{document}
\maketitle
\begin{abstract}
Although pre-training models have achieved great success in dialogue generation, their performance drops dramatically when the input contains an entity that does not appear in pre-training and fine-tuning datasets (unseen entity). To address this issue, existing methods leverage an external knowledge base to generate appropriate responses. In real-world scenario, the entity may not be included by the knowledge base or suffer from the precision of knowledge retrieval. 
To deal with this problem, instead of introducing knowledge base as the input, we force the model to learn a better semantic representation by predicting the information in the knowledge base, only based on the input context. 
Specifically, with the help of a knowledge base, we introduce two auxiliary training objectives: 1) Interpret Masked Word, which conjectures the meaning of the masked entity given the context; 2) Hypernym Generation, which predicts the hypernym of the entity based on the context. Experiment results on two dialogue corpus verify the effectiveness of our methods under both knowledge available and unavailable settings.
\end{abstract}

\begin{figure}[t!]
    \centering
        \subfigure[Non-knowledge dialogue generation. \label{fig:1a}]{
    \includegraphics[width=0.5\textwidth]{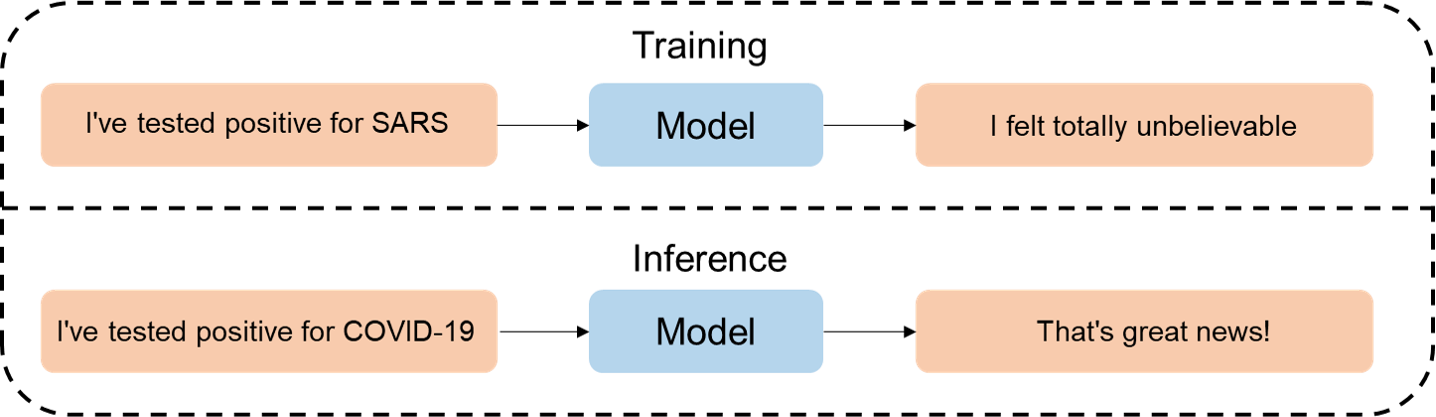}
    }
        \subfigure[Knowledge grounded dialogue generation. Note that the knowledge of ``COVID-19'' can not be retrieved from the knowledge base, because it is a new term. \label{fig:1b}]{
    \includegraphics[width=0.5\textwidth]{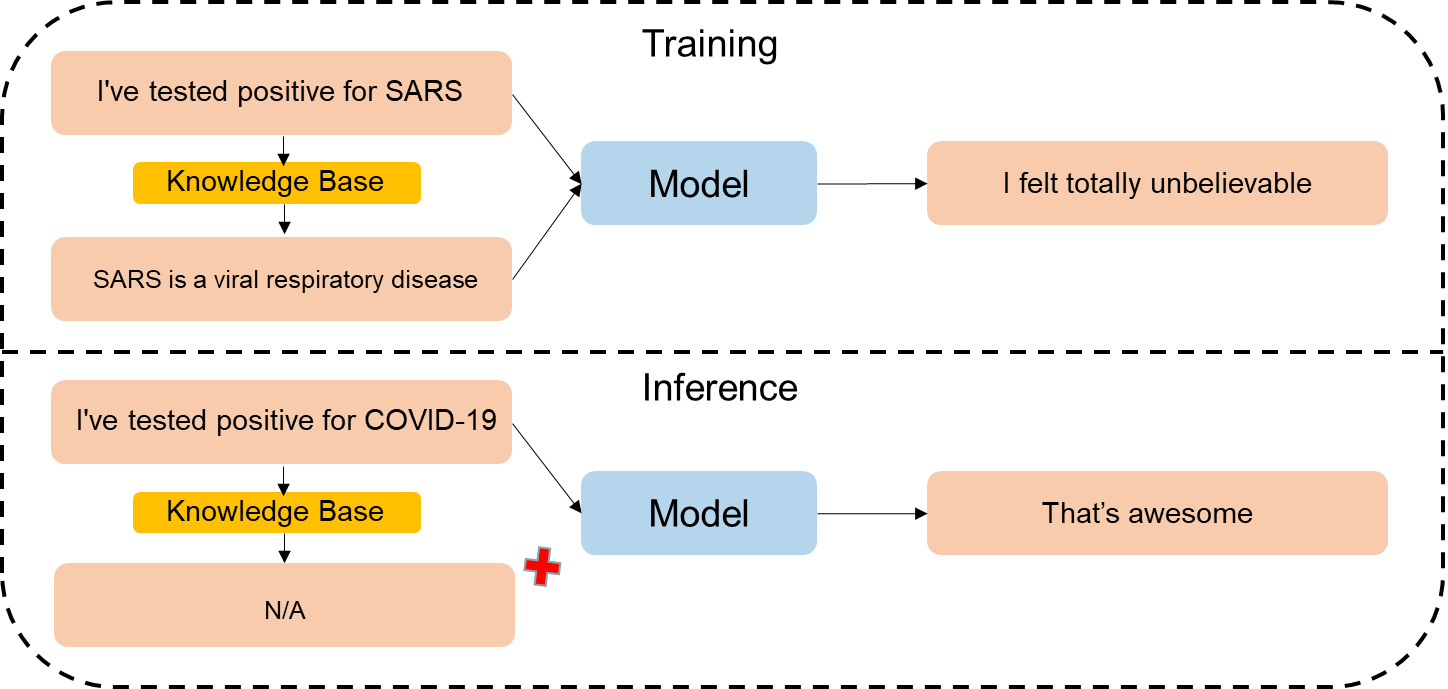}
    }
    \subfigure[The proposed knowledge enhanced dialogue generation. \label{fig:1c}]{
    \includegraphics[width=0.5\textwidth]{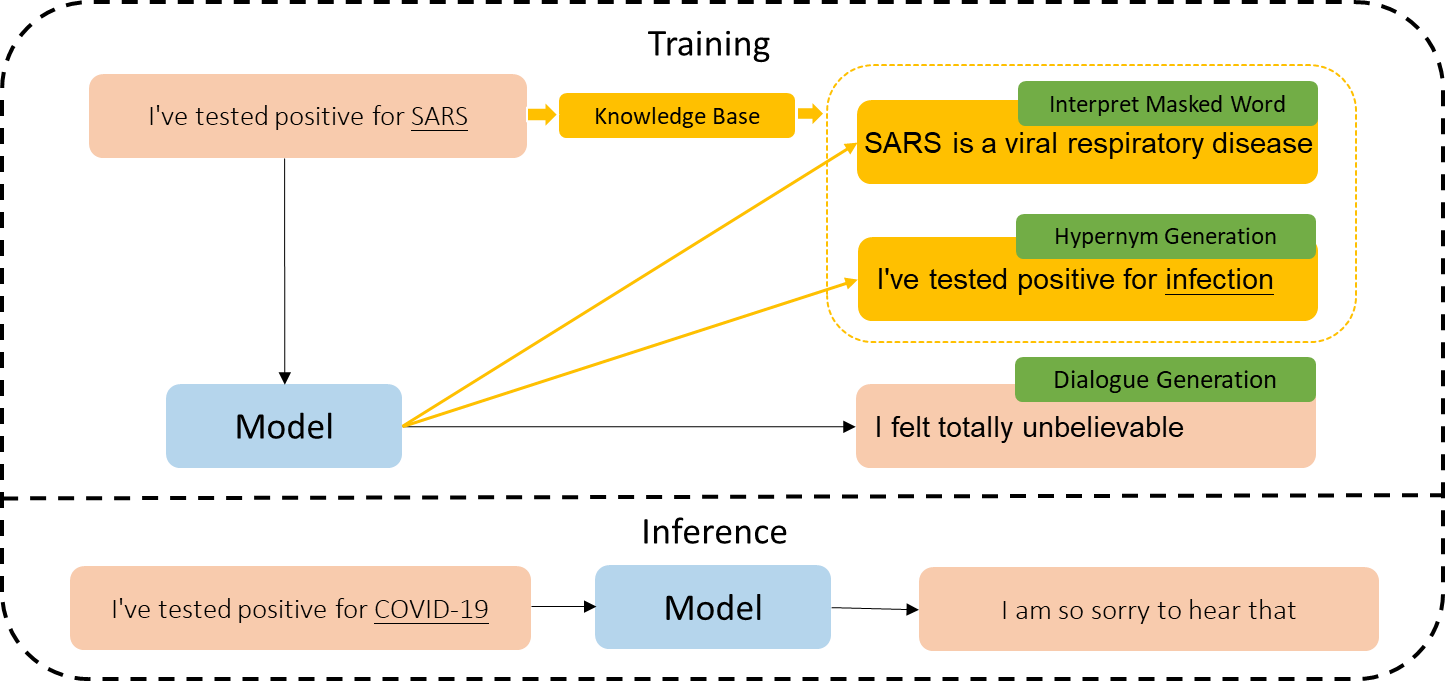}
    }
    \caption{An illustration of how knowledge can help dialogue generation for different methods.}
    \label{fig:intro}
\end{figure}

\section{Introduction}
Owing to large amounts of conversation data and pre-training models  \cite{dialogpt, blender}, generation-based chatbots have achieved significant advances and even reach human parity on specific testsets \cite{persona,wizard, bst}. 
However, the robustness of the pre-trained model is still low with regard to unseen entities \cite{zhang2016understanding, wizard}.
In practice, users often talk with chatbots about latest news and the recently hot topics \cite{blind-users}, which may not appear in the pre-training or fine-tuning corpus. 
For instance, ``COVID-19'' is a new term, which does not appear in the training data of Blender \footnote{Blender uses 1.5B Reddit 2019 text data for pre-training.} \cite{blender}, leading to poor performance when a user mentions ``COVID-19''. 
As shown in Figure~\ref{fig:1a}, given an utterance ``I've tested positive for COVID-19'', Blender yields a bad response ``That's great news'' because it misunderstands the utterance by the word ``positive'', which poses a real challenge for building an accurate and robust generation-based chatbot.




Existing methods leverage external knowledge to tackle the problem \cite{knowledge-grounded}, where chatbots retrieve relevant knowledge about the entities from a external knowledge base and use the retrieved knowledge to help generating appropriated responses. However, these methods heavily depend on the coverage of the knowledge base and the accuracy of knowledge retrieval, which may fail when the entity is not included by the knowledge base \cite{tnf} or the retrieved knowledge is inappropriate \cite{lian2019learning}. 
As shown in Figure~\ref{fig:1b}, the knowledge retriever fails to retrieve ``COVID-19'' from the knowledge base, yielding an incorrect response.
According to our statistics, the knowledge retrieval failure is not rare in real practice. Taking Reddit as an example, we collect 407 dialogues over 40 topics on the Trendings panel and find that 24.8\% of the topic words are polysemous, indicating the probability of incorrect knowledge retrieval, and 47.9\% of topic words are not included by the Wikipedia. To date, there are few studies that have investigated how to build a dialogue generation model within which knowledge may be unavailable during inference. 



We solve this problem by proposing a knowledge enhanced fine-tuning method, trying to understand semantic information of entities based on the context. For example, given the sentence ``\textit{I want to submit a paper to EMNLP}", a person may not know what ``\textit{EMNLP}" is, but he/she can guess that it should be a conference or a journal, based on the context.  
Similarly, we aim to enhance the semantic representation of unseen entities by  guiding the model to learn the meaning of the words only based on the context information.

To achieve this, we take Blender~\cite{blender} as our backbone model, and propose two auxiliary training objectives (Figure~\ref{fig:1c}) in fine-tuning, dubbed as \textbf{K}nowledge \textbf{E}nhanced \textbf{Blender} (\textbf{KE-Blender}).
The first objective is {\it Interpret Masked Word}, which predicts the word's definition based on the context, where the definition is obtained from a knowledge base. The second is {\it Hypernym Generation}, which predicts the corresponding hypernym of the word given by WordNet. 
These two introduced training objectives force the model to learn semantic information from the external knowledge base during training, guessing the meaning of the unseen entity with its context, so as to better understand the input utterance and generate relevant responses during inference.  
Both training objectives do not require further human labeling,  which makes it possible for extending to large-scale pre-training.

Results on the Wizard of Wikipedia benchmark show that the proposed model brings performance improvement. 
The proposed method achieves 14.9 and 18.4 PPL on Wizard Test Unseen in the knowledge available setting and unavailable setting, respectively, which outperforms the Blender baselines (16.3 and 19.9 PPL).
To further verify the effectiveness of our method in real-world scenarios, we collect 407 dialogues on the Reddit \textit{Trendings} panel, demonstrating the effectiveness of the proposed method in practice. 
We release our code and dataset at \url{https://github.com/Nealcly/KE-Blender}.

\section{Related Work}
\subsection{Knowledge Enhanced Pre-training}
BAIDU-ERNIE~\cite{baidu-ernie} uses entity-level masking and phrase-level masking strategy to enhance knowledge into language model.
THU-ERNIE~\cite{thu-ernie} incorporates contextual representations with separate KG embeddings.
LUKE~\cite{luke} proposes an entity-aware self-attention to boost the performance of entity related tasks.
SenseBERT~\cite{sensebert} uses WordNet to infuse the lexical semantics knowledge into BERT. 
KnowBERT~\cite{KnowBert} incorporates knowledge base into BERT using the knowledge attention.
TNF~\cite{tnf} accelerates pre-training by taking notes for the rare words.
Compared with these methods, which enhances the pre-trained encoder by utilizing named entities or knowledge base, we inject knowledge to improve the generation ability of seq2seq models given the unseen word.

\subsection{Knowledge Grounded Dialogue Generation}
With advances in deep learning, pre-trained language models have shown promising results in dialogue generation \cite{bart,dialogpt, blender}. To equip the models with external knowledge, \citet{persona} first show that adding user profile information is able to produce a more consistent and engaging response.
\citet{wizard} propose a Transformer memory network to retrieve knowledge from Wikipedia. \citet{itdd} use two-step decoding, which first generate a response based on context, and then take the generated response and relative knowledge as input to generate a new response. \citet{skt} focus on knowledge selection in dialogue generation by utilizing a sequential latent variable model. \citet{KDBTS} further enhance the selection module with the posterior information. \citet{knowledgeGPT} use reinforcement learning to optimize knowledge selection with unlabeled data.
Different from their work, our KE-Blender does not take knowledge as input, because knowledge is only used to enhance our model during training.

\section{Method}
\subsection{Task}
Suppose that we have a training set $\mathbb{D}^S = \{\mathbf{U}^S_i, \mathbf{K}^S_i, \mathbf{R}^S_i\}|_{i=1}^{|L|}$, where $\mathbf{U}^S_i$, $\mathbf{K}^S_i$ and $\mathbf{R}^S_i$ are the dialogue context, the external knowledge retrieved from the knowledge base and the response, respectively. In addition to $\mathbb{D}^S$, we have a test dataset $\mathbb{D}^P=\{\mathbf{U}^P,\mathbf{R}^P\}$. Unlike $\mathbb{D}^S$, $\mathbb{D}^P$ does not contain external knowledge, because associated background knowledge for unseen word is difficult to obtain in real time during inference.
Our goal is to learn a dialogue generation model $P(\mathbf{R}|\mathbf{U};\theta)$ with the help of $\mathbb{K}^S$, where $\theta$ is the parameters of the model.
It should be noted that, the dialogue generation model $P(\mathbf{R}|\mathbf{U};\theta)$ generates the response $\mathbf{R}$ only based on the input context $\mathbf{U}$, without using knowledge $\mathbf{K}$ as input. 

In the following sections, we will introduce the model structure first, and then show how to leverage the external knowledge $\mathbf{K}$ to enhance the generation model $P(\mathbf{R}|\mathbf{U};\theta)$ with our two proposed training objectives.

\begin{figure*}
    \centering
    \includegraphics[width=\textwidth]{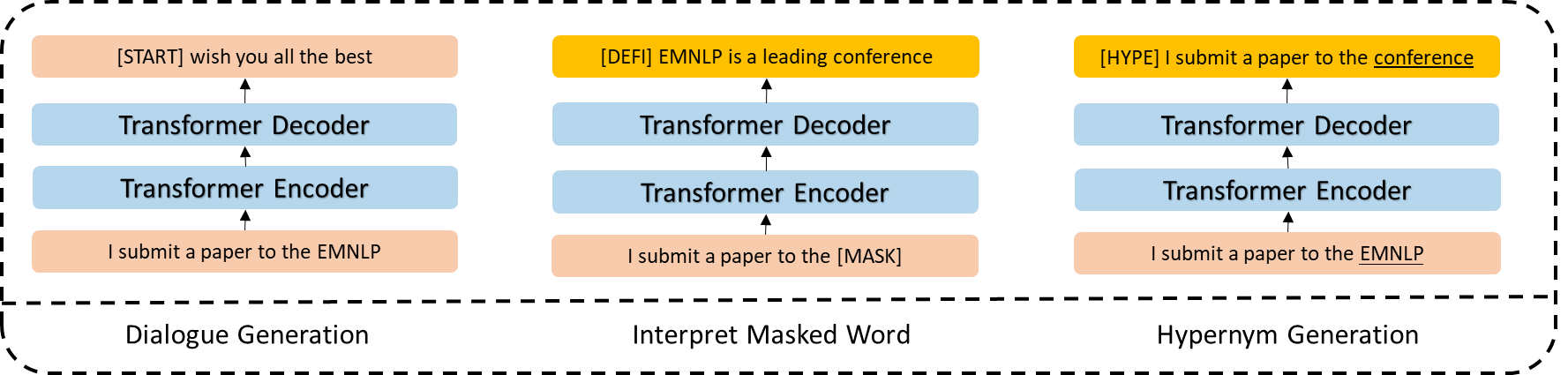}
    \caption{An illustration of three training objectives in KE-Blender. {\tt [START]}, {\tt [MASK]}, {\tt [DEFI]} and {\tt [HYPE]} are special tokens for KE-Blender. }
    \label{fig:method}
\end{figure*}

\subsection{Baseline}
\label{sec:baseline}
We consider Blender and {\bf K}nowledge {\bf G}rounded {\bf Blender} (KG-Blender) as our baselines in knowledge available and knowledge unavailable settings, respectively.
\paragraph{Blender}
Given a dialogue context $\mathbf{U} =\{\mathbf{u}_1,...,\mathbf{u}_{l-1}\}$, we first concatenate $\mathbf{U}$ as a sequence of sentences $\mathbf{U} =
\{x_1, x_2,\dots, x_T\}$. The response is denoted as $\mathbf{R} = \{y_1, y_2,\dots, y_{T'}\}$. We train our model on the basis of Blender, which is a standard Seq2Seq Transformer architecture with pre-training. 
In particular, we feed the dialogue context $\mathbf{U}$ to the encoder of the transformer, and then we obtain hidden representations of the sentence
\begin{equation}
\mathbf{h}^{enc} = \textsc{Transformer\_Encoder}(\mathbf{U}) 
\label{eq:encoder}
\end{equation}

At the $t$ th step of the decoder, $\mathbf{h}^{enc}$ and previous output tokens $y_{1:t-1}$ are then as inputs, yielding a representation using attention \cite{transforms}
\begin{equation}
\mathbf{h}^{dec}_t = \textsc{Transformer\_Decoder}(\mathbf{h}^{enc}, y_{1:t-1}) 
\end{equation} 

The generative probability distribution of $y_t$ is given by
\begin{equation}
    p(y_t|\mathbf{U},y_{1:t-1}) = softmax (\mathbf{W}^o \mathbf{h}^{dec}_t + \mathbf{b}^o)
\end{equation}
where $\mathbf{W}^o$ and $\mathbf{b}^o$ are trainable parameters.

We use the standard Maximum Likelihood Estimation to optimize the model parameters $\theta$. Given a training pair $(\mathbf{U},\mathbf{R})$, we minimize:
\begin{equation}
    \mathcal{L}_{dialogue} = - \sum_{t=1}^{T'} \log p(y_t|\mathbf{U},y_{1:t-1})
\label{eq:dialogue_loss}
\end{equation}

We adopt Blender-90M \cite{blender} to initialize our Seq2Seq Transformer model, which has been pre-trained on 1.5B training examples from Reddit 2019. 

\paragraph{Knowledge Grounded Blender}
One intuitive baseline to use knowledge is to take both the context and the knowledge as input.
In particular, the concatenation of the context $\mathbf{U}$ and the associated knowledge $\mathbf{K}$ is fed to the transformer encoder:
\begin{equation}
\begin{split}
&\mathbf{h}^{enc'} = \textsc{Transformer\_Encoder}([\mathbf{U};\mathbf{K}])  \\
&\mathbf{h}^{dec'}_t = \textsc{Transformer\_Decoder}(\mathbf{h}^{enc'}, y_{1:t-1}) \\ 
    & p(y_t|\mathbf{U},\mathbf{K},y_{1:t-1}) = softmax (\mathbf{W}^o \mathbf{h}^{dec'}_t + \mathbf{b}^o)
\end{split}
\end{equation}

Similar to Eq~\ref{eq:dialogue_loss}, given a training pair $(\mathbf{U},\mathbf{K},\mathbf{R})$, the loss function is
\begin{equation}
    \mathcal{L}_{dial\_know} = - \sum_{t=1}^{T'} \log p(y_t|\mathbf{U},\mathbf{K},y_{1:t-1})
\end{equation}

Note that it is difficult to use KG-Blender directly when knowledge is unavailable, because KG-Blender relies knowledge as input. 

\subsection{Knowledge Enhanced Blender}
To build a robust model for knowledge unavailable setting, we consider adding two auxiliary loss during fine-tuning.
People try to understand an unseen word based on the context, even if a dictionary is unavailable. To simulate this behavior, we explicitly guide the Blender model to learn the meaning of words only based on the context information.


\paragraph{Interpret Masked Word}
The first objective is to ask the model to restore the definition of masked words. We can use different methods to select which words should be masked. For example, we could mask proper nouns in the utterance, or pre-defined topic word for specific dataset\footnote{Wizard of Wikipedia dataset have defined topic words for each dialogue.}. For example, the input text is \textit{``I submit a paper to the EMNLP''}. ``\textit{EMNLP}'' is replaced by {\tt [MASK]}, yielding \textit{``I submit a paper to the {\tt [MASK]}''}. The definition retrieved from Wikipedia is ``\textit{EMNLP is a leading conference in the area of natural language processing and artificial intelligence }''. Then, the pre-trained model is required to restore the definition by consuming the masked utterance as input. In this way, the model is explicitly guided to understand the background knowledge of the masked word given the context. 

Formally speaking, given a single utterance $\mathbf{u}_{l-1} =
\{x_1, x_2,\dots, x_T\}$, we assume that $x_i$ is the topic word in $\mathbf{u}_{l-1}$, and its corresponding definition is denoted as $\mathbf{K}_{x_i}=\{k_1,k_2,\dots,k_{|K_{x_i}|}\}$. We use the special token {\tt [MASK]} to replace $x_i$ yielding $\mathbf{u}'_{l-1} = \{x_1,\dots,x_{i-1},${\tt [MASK]}$, x_{i+1},\dots,x_T'\}$ as the input of Eq~\ref{eq:encoder} in Section~\ref{sec:baseline}.  
To distinguish with the original dialogue generation task, we use a specific start token {\tt [DEFI]} to mark that the target sequence is the definition.
Given a training pair $(\mathbf{u}'_{l-1},\mathbf{K}_{x_i})$, the training objective of the interpret masked word is:
\begin{equation}
    \mathcal{L}_{interpret} = - \sum_{t=1}^{|K_{x_i}|} \log p(k_t|\mathbf{u}'_{l-1},k_{1:t-1})
\end{equation}

\paragraph{Hypernym Generation}
We also reconstruct the input utterance by replacing the topic words with the corresponding hypernym. Compared with topic words, the semantic field of its hypernym is more general. We use WordNet to construct our training instances. For instance, given an utterance $\mathbf{u}_{l-1}=$ \{\textit{I submit a paper to the EMNLP}\}, we use ``\textit{conference}'' to replace ``\textit{EMNLP}'', where ``\textit{conference}'' is the hypernym of ``\textit{EMNLP}'', yielding the target sequence $\mathbf{u}''_{l-1}=$ \{\textit{I submit a paper to the conference}\}. This training objective aims to guide the model to understand the semantic information of unseen words. We use a specific start token {\tt [HYPE]} to mark the target sequence is the hypernym generation. Given a training pair $(\mathbf{u}_{l-1},\mathbf{u}_{l-1}'')$, the training objective of the hypernym generation is:
\begin{equation}
    \mathcal{L}_{hypernym} = - \sum_{t=1}^{|u''_{l-1}|} \log p(y''_t|u_{l-1},y''_{1:t-1})
\end{equation}

\paragraph{Training}
We optimize the dialogue generation loss with the two external loss at the same time:
\begin{equation}
\small
    \mathcal{L} = \sum_{i=1}^{|L|} \mathcal{L}^i_{dialogue} +
    \sum_{i=1}^{|K|} \mathcal{L}^i_{interpret} + \sum_{i=1}^{|H|} \mathcal{L}^i_{hpernym}
\end{equation}
where $|K|$ and $|H|$ represent the number of training instances for {\it Interpret Masked Word} and {\it Hypernym Generation}, respectively.

\paragraph{Inference}
During inference, we only take the context as input if the additional retrieved knowledge is unavailable. Following~\citet{knowledgeGPT}, we adopt greedy search to select the highest probability token at each time step. We denote our model as {\bf K}nowledge {\bf E}nhanced {\bf Blender} (KE-Blender) for the remaining of this paper.


\section{Experiments}
\subsection{Datasets}
{\bf Wizard of Wikipedia} \cite{wizard} is a knowledge aware chit-chat dialogue benchmark, where each instance has an initial topic given by two annotators. 
The dataset contains 18,430 training dialogues with 1,365 topics, and each topic is linked to a Wikipedia article. 
Its test set is split into Test Seen and Test Unseen based on whether the topic is appear in the training set. We evaluate our methods with several baselines on both Test Seen and Test Unseen.

There are 148,357 training instances in the Wizard of Wikipedia training set. To enhance knowledge into the model, we further construct 65,072 training pairs for interpret masked word based on Wikipeida and 86,612 training pairs for hypernym generation based on WordNet.


{\bf Reddit Trendings} is a test set to simulate real-world settings, by crawling users' dialogue from its Trendings panel in 2021. Reddit Trendings panel contains the latest hot topics, and most of them are not included in the external knowledge bases. We first obtain topic words from the Reddit Trendings panel, then crawl the dialogue based on the topic words.
We further filter the datasets by selecting out dialogue that includes at least 2 utterances, yielding a dataset which similar to the Wizard setting.
Finally, the dataset consists of 407 utterances over 40 trending topics. 
\subsection{Setup}
We implement KE-Blender with {\tt transformers} and choose {\tt blenderbot-90M} as the pre-trained language model. AdamW with a batch size of 128 is used to optimize parameters. The initial learning rate is set as 1e-5, which is halved in each training iteration. 
We set the maximum input tokens as 512. To ensure that KE-Blender also works well in knowledge available settings, we also create extra training instances by concatenating the context with the associated knowledge as input.
\subsection{Baselines}
We compare KE-Blender with Blender and KG-Blender, also drawing the following state-of-the-art methods as reference:

{\bf Transformer} \cite{transforms} is a standard transformer model for dialogue generation. It takes the concatenation of context utterances and the associated knowledge as input.

{\bf SKT} \cite{skt} uses a sequential latent variable model for knowledge selection, and then generates the response based on the context and the selected knowledge. 

{\bf DRD} \cite{drd} is a pre-training model designed for the low-resource dialogue generation, which decomposes the decoder into
independent components.

{\bf SKT + GPT-2} \cite{knowledgeGPT} feeds the knowledge selected by SKT to GPT-2 for dialogue response generation.

{\bf SKT + PIPM + KDBTS} \cite{KDBTS} uses posterior information to help prior knowledge selection module, and trains the decoder with knowledge distillation.

{\bf KnowledGPT} \cite{knowledgeGPT} adopts reinforcement learning to optimize the knowledge selection module, which gives state-of-the-art performance on Wizard. 

{\bf Blender-FT.} Blender is a large-scale dialogue pre-training model. We fine-tune the Blender on Wizard training set without utilizing external knowledge.

{\bf KG-Blender.} We fine-tune Blender on the Wizard training set by concatenating the context and the associated knowledge as the input. In the setting where external knowledge is unavailable, only context is used to generate response.

\begin{table*}[t!]
    \centering
    \small
    \begin{tabular}{c|c|c|c|c|c|c}
    \toprule
         \multirow{2}{*}{{\bf Model}} & \multicolumn{2}{c|}{{\bf Test Seen}} & \multicolumn{2}{c|}{{\bf Test Unseen}} & \multicolumn{2}{c}{{\bf Performance Gap}} \\
    \cmidrule(r){2-7}
          & {\bf PPL} & {\bf F1} & {\bf PPL} & {\bf F1} & {\bf PPL} & {\bf F1} \\
    \midrule
            \multicolumn{7}{c}{\bf w/ knowledge during inference} \\
    \midrule
         Transformer MemNet \cite{wizard} & 66.5 & 15.9 & 103.6 & 14.3 & 37.1 & 1.6 \\
         SKT \cite{skt} & 52.0 & 19.3 & 81.4 & 16.1 & 34.6 & 3.2 \\
         DRD \cite{drd} & 19.4 & 19.3 & 23.0 & 17.9 & 3.6 & 1.4 \\
         SKT + GPT-2 \cite{knowledgeGPT} & 17.6 & 20.3 & 23.7 & 17.8 & 6.1 & 2.5 \\
         SKT+PIPM+KDBTS \citet{KDBTS} & 42.7 & 19.9 & 65.7 & 17.6 & 23.0 & 2.3 \\
         KnowledGPT \cite{knowledgeGPT} & 19.2 & {\bf 22.0} & 22.3 & {\bf 20.5} & 3.1 & 1.5 \\
    \midrule
         KG-Blender $\dagger$ & 13.8 & 18.4 & 16.3 & 17.8 & 2.5 & 0.6 \\
         KE-Blender (Ours) & {\bf 13.4} & 18.1 & {\bf 14.9} & 17.6 & {\bf 1.5} & {\bf 0.5} \\
    \midrule
    \multicolumn{7}{c}{\bf w/o knowledge during inference} \\
    \midrule
        Repeat last utterance & - & 13.8 & - & 13.7 & - & - \\
    \midrule
        Blender-FT $\dagger$ & 16.1 & 16.5 & 19.9 & 12.9 & 3.8 & 3.6\\
        KG-Blender $\dagger$ & 18.6 & 15.5 & 22.7 & 14.7 & 4.1 & 0.7 \\
        KE-Blender (Ours) & {\bf 15.5} & {\bf 17.0} & {\bf 18.4} & {\bf 16.7} & {\bf 2.9} & {\bf 0.3} \\
    \bottomrule
    \end{tabular}
    \caption{Performance on Wizard Test Seen and Wizard Test Unseen. Note that the lower PPL and the higher F1 indicate better generation model. ``Performance Gap'' represents the performance gap between Test Seen and Test Unseen, the lower Performance Gap indicates better generalization ability. $\dagger$ indicates our baseline implementation.}
    \label{tab:wizard}
\end{table*}

\begin{table}[t!]
    \centering
    \small
    \begin{tabular}{c|c|c|c|c}
    \toprule
         Model & R@1 & R@5 & R@10 & R@20 \\
    \midrule
         Human Reference & 41.59 & 60.96 & 65.95 & 70.33 \\
    \midrule
            \multicolumn{5}{c}{w/ knowledge during inference} \\
    \midrule
         KG-Blender & 37.41 & 57.29 & 62.48 & 66.56 \\
         KE-Blender & {\bf 39.14} & {\bf 58.72} & {\bf 62.70} & {\bf 67.18} \\
    \midrule
    \multicolumn{5}{c}{w/o knowledge during inference} \\
    \midrule
        Blender & 31.91 & 49.34 & 56.07 & 60.75\\
        KG-Blender & 31.89 & 49.21 & 56.03 & 60.78 \\
        KE-Blender & {\bf 32.11} & {\bf 51.58} & {\bf 57.29} & {\bf 63.03} \\
    \bottomrule
    \end{tabular}
    \caption{Knowledge performance based on BERT-large on Wizard Test Unseen.}
    \label{tab:wizard-knowledge}
\end{table}

\begin{table}[t!]
    \centering
    \small
    \begin{tabular}{c|c|c|c|c}
    \toprule
         Model & Fluency & Know & Coherence & Kappa \\
    \midrule
            \multicolumn{5}{c}{w/ knowledge during inference} \\
    \midrule
         KG-Blender & {\bf 1.94} & 1.63 & 1.70 & 0.63 \\
         KE-Blender & 1.93 & {\bf 1.65} & {\bf 1.74} & 0.64 \\
    \midrule
    \multicolumn{5}{c}{w/o knowledge during inference} \\
    \midrule
        Blender & {\bf 1.95} & 1.37 & 1.51 & 0.68\\
        KG-Blender & 1.89 & 1.43 & 1.47 & 0.59 \\
        KE-Blender & 1.92 & {\bf 1.62} & {\bf 1.70} & 0.65 \\
    \bottomrule
    \end{tabular}
    \caption{Human Evaluation on Wizard Test Unseen. Know-Knowledgeable.}
    \label{tab:wizard-human}
\end{table}

\subsection{Metrics}

\paragraph{Automatic evaluation metrics:} Following \citet{wizard} and \citet{skt}, models are measured using the perplexity of the ground-truth response (PPL) and unigram F1-score (F1). 

\begin{figure}[t!]
    \centering
    \includegraphics[width=0.5\textwidth]{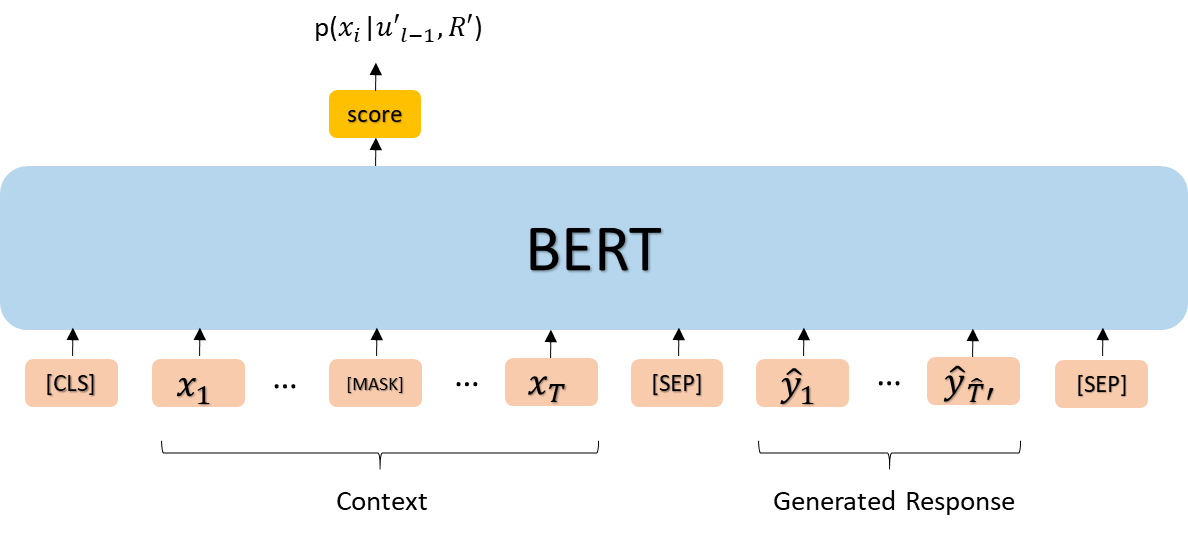}
    \caption{The proposed BERT evaluation method.}
    \label{fig:eval_bert}
\end{figure}

\citet{bert-evaluate} show that BERT can be used to evaluate the generated response. We employ BERT-based evaluation metrics to evaluate whether the generated response is knowledgeable as supplements to PPL and F1. As shown in Table~\ref{fig:eval_bert}, the dialogue generation model is first required to generate response $\hat{\mathbf{R}}$ based on the dialogue context $\mathbf{U}=\{\mathbf{u}_1,\dots,\mathbf{u}_{l-1}\}$. We use the special token {\tt [MASK]} to replace the topic word in the last context utterance $\mathbf{u}_{l-1}$. Then a masked language model (i.e. BERT-large) is used to predict the masked topic word using the last context utterance $\mathbf{u}_{l-1}$ and the generated response $\hat{R}$. The recall$@k$ for the masked language model is used to measure the knowledge stored in the dialogue generation model.
Intuitively, if a dialogue generation model is more knowledgeable, the masked language model is stronger to predict the masked topic word based on the generated response $\hat{\mathbf{R}}$ and last context utterance $\mathbf{u}_{l-1}$. 

\paragraph{Human evaluation metrics:}
Manual evaluations are essential for evaluating dialogue generation \cite{human-eval}. We conduct human evaluations to compare KE-Blender with our baseline Blender and KG-Blender by randomly sampling 200 instances from the Wizard Test Unseen. We define three metrics for manual evaluation, including fluency, knowledgeability and coherence. Each aspect is scored into three grades, 0, 1 and 2, representing “bad”, “normal” and “good” respectively. Following~\citet{wu2018response}, we employ three 
annotators to do a side-by-side human evaluation, and report the Fleiss Kappa \cite{fleiss1971mns} to show the agreement among human annotators.
\subsection{Results}
Table~\ref{tab:wizard} reports automatic results on the Wizard of Wikipedia dataset. 

\paragraph{Test Seen vs Test Unseen}
Compared with Test Seen, all models perform worse on Test Unseen, especially where knowledge is unavailable. For example, the Blender-FT only achieves F1 scores of 16.5 and 12.9 on Test Seen and Test Unseen, respectively. Compared with several baselines, KE-Blender gives the lowest performance gap, suggesting that KE-Blender is more robust when it comes to Test Unseen.

\paragraph{w/ Knowledge During Inference}

Previous work \cite{skt, knowledgeGPT} focuses on how to better leverage knowledge for dialogue generation.
Compared with these strong baselines, KE-Blender performs competitively in the knowledge-available setting, even though KE-Blender is designed for knowledge-unavailable setting. Notably, it achieves the best reported PPL on Wizard.
The results are consistent with our intuition. Knowledge grounded methods perform well when knowledge is provided during inference, and our method is robust and does not degrade in the w/ knowledge setting.


\begin{table}[t!]
    \centering
    \small
    \begin{tabular}{c|c|c}
    \toprule
        Model & PPL & F1  \\
    \midrule
        KE-Blender & {\bf 18.36} & {\bf 16.73} \\
        w/o Interpret & 18.75 & 16.29\\
        w/o Hypernym & 18.95 & 16.27 \\
        Blender  & 19.87 & 12.91 \\
    \bottomrule
    \end{tabular}
    \caption{Ablation study on the Wizard Test Unseen when knowledge unavailable.}
    \label{tab:ablation}
    \vspace{-0.5cm}
\end{table}

\begin{figure}
    \centering
    \includegraphics[width=0.5\textwidth]{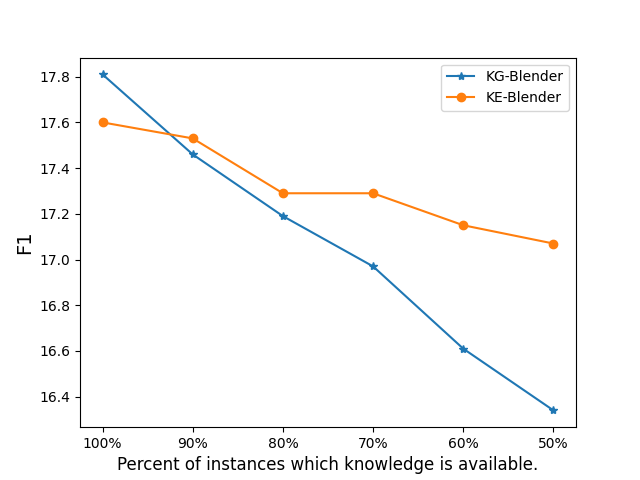}
    \caption{Model performance when part of knowledge is available during inference. 90\% indicates the 90\% of instances are knowledge available.}
    \label{fig:percent}
\end{figure}

\begin{table*}[t!]
    \centering
    \small
    \begin{tabular}{c|c|c|c|c|c|c|c|c|c|c}
    \toprule
          \multirow{2}{*}{Setting} & \multirow{2}{*}{\%} & \multicolumn{3}{c|}{\bf w/ knowledge} & \multicolumn{3}{c|}{\bf w/o knowledge} & \multicolumn{3}{c}{{\bf Performance Gap}} \\
    \cmidrule{3-11}
         && KE & KG & Blender & KE & KG & Blender & KE & KG & Blender \\
    \midrule
         None & 47.9 & {\bf 57.5} & 62.2 & 90.3 & {\bf 57.5} & 62.2 & 90.3 & \ \ 0.0 &\ \ 0.0 & \ \ 0.0 \\
         Invalid & 24.8 & {\bf 58.2} & 66.8 & 80.1 & {\bf 51.1} & 56.3 & 90.6 & -7.1 & -10.5 & -10.5 \\
         Valid & 27.3 & {\bf 53.8} & 57.6 & 93.6 & {\bf 53.5} & 56.9 & 88.0 & \ \  0.3 & \ \ 0.7 & -13.6 \\
         Overall & 100 & {\bf 56.6} & 61.8 & 88.3 & {\bf 54.7} & 59.3 & 89.4 &\ \ 1.9 &\ \ 2.5 & 1.3 \\
    \bottomrule
    \end{tabular}
    \caption{Model Performance (PPL) on Reddit Trendings. KE - KE-Blender, KG - KG-Blender. ``None'', ``Invalid'' and ``Valid'' are three subset of the test set, which indicates knowledge can not be retrieved, knowledge can be retrieved but it is incorrect, and gold knowledge is available, respectively. \% represents the percent of instances on test set. ``Performance Gap'' represents the performance gap between w/ knowledge and w/o knowledge. The ``Performance Gap'' close to zero indicates the model does not depend on external knowledge.}
    \label{tab:reddit}
\end{table*}

\vspace{-0.2cm}
\paragraph{w/o Knowledge During Inference}
When external knowledge is unavailable during the inference stage, knowledge grounded methods cannot be directly applied to this setting since it requires knowledge as an input. Hence, we compare KE-Blender with Blender and KG-Blender. As can be seen from Table~\ref{tab:wizard}, our method shows large advantages in all metrics, achieving a 16.7 F1 score. It shows that our external training objectives can help model to generalize better when meet unseen words.

\paragraph{BERT Evaluation}
Table~\ref{tab:wizard-knowledge} shows the recall of masked topic words predicted by a BERT model, where a higher recall score indicates the stronger correlation between the knowledge and the response. Human's response obtains a higher score, which means our evaluation metric is reasonable and there is still a gap between human's  reply and machine's reply.
Our method gives the best performance in both settings, demonstrating strong performance when knowledge is absent, which shows that our auxiliary training objectives is able to help model to learn a better semantic representation. 
Surprisingly, it also outperforms simple knowledge grounded methods when knowledge is available. 


\paragraph{Human Evaluation}
Table~\ref{tab:wizard-human} compares KE-Blender with baselines using human evaluation. All models are able to produce fluent response due to the power of pre-training. Inference with the retrieved knowledge is particularly helpful for the model to generate a more knowledgeable and coherent response. 
When knowledge is unavailable, KE-Blender significantly outperforms Blender and KG-Blender ($p$\textless0.01) measured in both knowledgeable and coherent, also giving highly competitive results with the model using knowledge as input. The value of Fleiss’ Kappa \cite{fleiss1971mns} exceed 0.59 on all models, showing a high inter-rater agreement among annotators.



\begin{table*}[t!]
    \centering
    \small
    \begin{tabular}{c|c}
    \toprule
    \multicolumn{2}{c}{\bf Dialogue Generation} \\
    \midrule
        Speaker A & Hey! I enjoy Elvis music, do you like his music? \\
        Speaker B & Yes, I love to listen his music. He was not just a singer, he was an actor too. \\
        Speaker A & Oh, I didn't know that! Do you know any movies he acted in? \\
    \midrule
    \multicolumn{2}{c}{\bf w/o knowledge during inference} \\
    \midrule
        Blender-FT & Oh wow, that's a long time ago! I'm sure it's been around for a long while. \\
        KE-Blender & I know he was in a lot of movies, he was also in a few TV shows. \\
    \midrule
    \multicolumn{2}{c}{\bf w/ knowledge during inference} \\
    \midrule
    \multirow{2}{*}{Knowledge} & Elvis Aaron Presley was an American singer, musician, and actor. Jailhouse Rock is a American musical \\ & drama film directed by Richard Thorpe and starring Elvis Presley, Judy Tyler, and Mickey Shaughnessy. \\
    \midrule
    KG-Blender & He was in a movie called jailhouse rock. It was about a young man who was mentored in music by a prison cellmate. \\
    KG-Blender & He was in a movie called jailhouse rock. It was a really good movie. \\
    \bottomrule
    \end{tabular}
    \caption{Examples of generated responses on the Wizard of Wikipedia Test Unseen.}
    \label{tab:case1}
\end{table*}


\begin{table}[t!]
    \centering
    \small
    \begin{tabular}{c|c}
    \toprule
    \multicolumn{2}{c}{Interpret Masked Word} \\
    \midrule
    \multirow{2}{*}{In} & I never realized that Instagram is  \\
    & owned by {\tt [MASK]}, did you? \\
    \midrule
    \multirow{2}{*}{Out} & Facebook is an American online social media \\
    & and social networking service company. \\
    \midrule
    In & I really like {\tt [MASK]}. But blue is also a nice color. \\
    Out & Purple is a color intermediate between blue and red. \\
    \midrule
    \multicolumn{2}{c}{Hypernym Generation} \\
    \midrule
        In & What else you know about \underline{bowling}? \\
        Out & What else you know about \underline{ball game}? \\
    \midrule 
        In & I'm sorry to hear that, I have no \underline{pets}. \\
        Out & I'm sorry to hear that, I have no \underline{animals}. \\
    \bottomrule
    \end{tabular}
    \caption{Examples of generated definition and hypernym on the Wizard of Wikipedia Test Unseen. The knowledge does not exist in the training set. ``In'' and ``Out'' denote the input and output of KE-Blender, respectively.}
    \label{tab:case3}
\end{table}

\paragraph{Low-Knowledge-Resource Setting}
To simulate a low-knowledge-resource setting, we start from using the full knowledge in Wizard Test Unseen, and gradually reduce the amount of knowledge by randomly removing some entries. Figure~\ref{fig:percent} shows the trends when different percentage of knowledge in Test Unseen is removed. As the ablation knowledge increases, the performance of the two methods significantly decreases. The F1 of KG-Blender sharply decreases from 17.8 to 16.3. Compared with KG-Blender, the rate of decrease is much smaller for KE-Blender, which shows the effectiveness of knowledge enhanced fine-tuning.

\paragraph{Reddits Trendings}
We train KE-Blender and KG-Blender on Wizard training set, and test on Reddits Trendings, taking off-the-shelf Blender as a reference. The results are reported in Table~\ref{tab:reddit}. Note that there is a domain gap between Wizard and Reddits, which leads to a worse performance on Reddits Trendings.
KE-Blender achieves 56.6 and 54.7 PPL on w/knowledge and w/o knowledge settings, respectively, outperforming KG-Blender and off-the-shelf Blender in all settings.
When invalid knowledge is used as input, KG-Blender achieves 66.8 PPL in w/knowledge setting, which underperforms w/o knowledge setting (56.3 PPL).
This shows that the inappropriate knowledge selection has a destructive impact on models \cite{lian2019learning}. Integrating with valid knowledge, both models are able to generate more informative responses. Furthermore, KE-Blender gets the best performance gap, which confirms that KE-Blender does not rely on external knowledge base, demonstrating the effectiveness of the proposed auxiliary training objectives.

\section{Analysis}

\paragraph{Ablation Study} An interesting question is to explore the contribution of the two auxiliary losses in training. The results are shown in Table~\ref{tab:ablation}. We can see that each loss contributes a lot in automatic evaluation, with F1 increasing largely by adding each objective. When combining the two losses, there is still an improvement but marginal, which indicates the two loss may play similar roles for pre-training model. 

\paragraph{Case Study}
Table~\ref{tab:case1} shows an example of the model responses on the Wizard of Wikipedia Test Unseen. Under the knowledge-available setting, all models generate reasonable responses with the help of relevant knowledge. Both models mention that ``Elvis'' was in ``jailhouse rock'' by consulting the external knowledge base. When knowledge is unavailable, Blender-FT gives a non-informative response because it cannot understand the word ``Elvis''. In contrast, KE-Blender shows superior performance by producing informative and knowledgeable responses, which directly points out that ``Elvis'' appears in a lot of movies and also in a few TV shows. This case shows that our model can significantly improves response quality when knowledge is absent, while sustain good performance when knowledge is available.

\paragraph{Knowledge Mining} Although we add two additional tasks in training, it is unclear how well the model performs in these two tasks. Therefore, we further evaluate whether explicit knowledge can be recovered from our model given the unseen entity. First, we find that the perplexity of the ground-truth Wikipedia knowledge on Test Unseen is only 6.81. As shown in Table~\ref{tab:case3}, our model is able to produce reasonable definition based on context information and the pre-trained knowledge, and generate hypernyms for a given word associated with context.
These show that rich-knowledge is stored in KE-Blender during knowledge enhanced fine-tuning, which potentially allows us to ground open domain dialogues without external knowledge.

\section{Conclusion}
We presented KE-Blender for better handling response generation based on unseen words, which enables a model to generate knowledgeable response without external knowledge during inference. To explicitly inject the knowledge into the model, we proposed two training objectives, including interpret masked word and hypernym generation. To simulate real-world scenario, we also released a test set on Reddit Trendings.
Results on Wizard and Reddit Trendings show that KE-Blender outperforms several state-of-the-art methods and strong baselines in settings both when external knowledge is available and unavailable. 

\section*{Acknowledgments}
We would like to thank the anonymous reviewers for their
constructive comments, and Yulong Chen, Jingyi Liao and Sen Yang for insightful discussion and proofreading. Yue Zhang is the corresponding author.

\bibliography{anthology, custom}
\bibliographystyle{acl_natbib}

\end{document}